\documentclass[letterpaper]{article} 
\usepackage{aaai25}  
\usepackage{times}  
\usepackage{helvet}  
\usepackage{courier}  
\usepackage[hyphens]{url}  
\usepackage{graphicx} 
\usepackage{natbib}  
\usepackage{caption} 
\usepackage{algorithm}
\usepackage{algorithmic}

\usepackage{newfloat}
\usepackage{listings}

\usepackage{amsmath}
\usepackage{amsfonts}
\usepackage{multirow}

\usepackage{arydshln} 
\usepackage{lipsum} 

\usepackage[table]{xcolor} 
\usepackage{colortbl}       
\usepackage{booktabs}       
\usepackage{musixtex} 
\usepackage{tablefootnote}  
\usepackage{subcaption} 

\usepackage{fontawesome}
\definecolor{lightgray}{gray}{0.6}

\DeclareCaptionStyle{ruled}{labelfont=normalfont,labelsep=colon,strut=off} 
\floatstyle{ruled}
\newfloat{listing}{tb}{lst}{}
\floatname{listing}{Listing}
\usepackage{tikz} 

\title{\textit{VEGAS}: Towards Visually Explainable and Grounded Artificial Social Intelligence}

\author{
    Hao Li\textsuperscript{\rm 1}, 
 	Hao Fei\textsuperscript{\rm 2},
    Zechao Hu\textsuperscript{\rm 1},
    Zhengwei Yang\textsuperscript{\rm 1},
    Zheng Wang\textsuperscript{\rm 1}\thanks{Corresponding author}
}
\affiliations{
    \textsuperscript{\rm 1} National Engineering Research Center for Multimedia Software, 
    School of Computer Science, Wuhan University \\
   \textsuperscript{\rm 2} School of Computing, National University of Singapore

   \{sc.lihao, hzc\_aim, yzw\_aim, wangzwhu\}@whu.edu.cn, haofei37@nus.edu.sg
}

\usepackage{bibentry}

\begin{document}

\maketitle

\begin{abstract}
Social Intelligence Queries (Social-IQ) serve as the primary multimodal benchmark for evaluating a model's social intelligence level. While impressive multiple-choice question (MCQ) accuracy is achieved by current solutions, increasing evidence shows that they are largely, and in some cases entirely, dependent on language modality, overlooking visual context. Additionally, the closed-set nature further prevents the exploration of whether and to what extent the reasoning path behind selection is correct.
To address these limitations, we propose the Visually Explainable and Grounded Artificial Social Intelligence (\textbf{VEGAS}) model.  
As a generative multimodal model, VEGAS leverages open-ended answering to provide explainable responses, which enhances the clarity and evaluation of reasoning paths. 
To enable visually grounded answering, we propose a novel sampling strategy to provide the model with more relevant visual frames. We then enhance the model's interpretation of these frames through Generalist Instruction Fine-Tuning (GIFT), which aims to: i) learn multimodal-language transformations for fundamental emotional social traits, and ii) establish multimodal joint reasoning capabilities.  
Extensive experiments, comprising modality ablation, open-ended assessments, and supervised MCQ evaluations, consistently show that VEGAS effectively utilizes visual information in reasoning to produce correct and also credible answers. 
We expect this work to offer a new perspective on Social-IQ and advance the development of human-like social AI.
\end{abstract}
\begin{links}
\link{Code}{https://github.com/lihao921/VEGAS}
\end{links}

\begin{figure}[t]
	\centering 
	\includegraphics[width=0.46\textwidth]{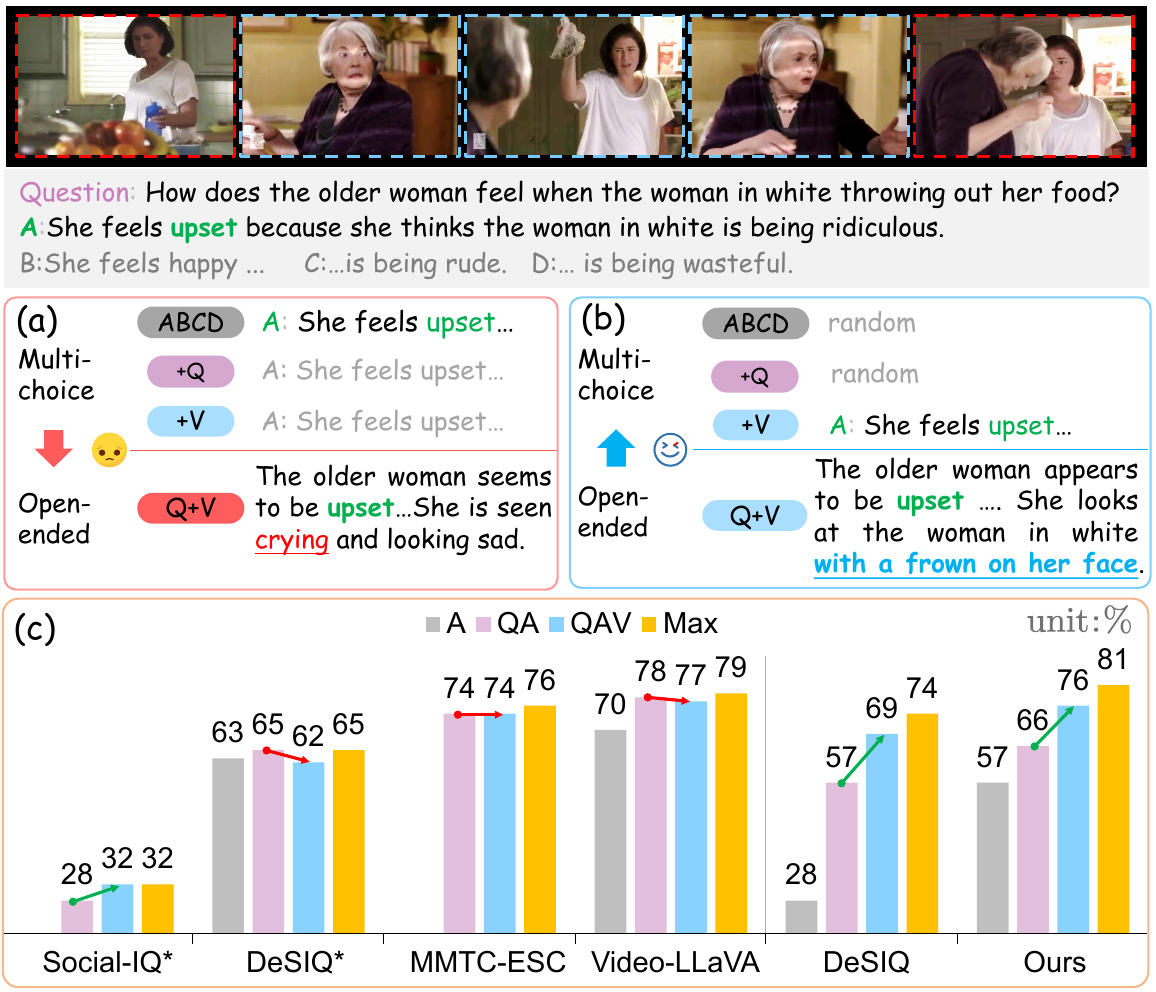}
	\caption{(a) An existing approach selects the correct option without knowing the question or even the video context, revealing incorrect rationale in the open-ended answers. (b) Our study begins with a correct reasoning path grounded in the video, ensuring reliable selection. (c) Our model enhances visual engagement, reduces the language shortcut, and achieves comparable but more reliable MCQ accuracy. * denotes the baseline of the corresponding method.  }
	\label{lead_in_fig} 
\end{figure}

\section{Introduction}
Human-level AI necessitates not only formidable reasoning abilities at the individual level, but also must exhibit social intelligence akin to that of humans in social contexts.
This motivates the study of the topic of Artificial Social Intelligence (ASI) \cite{bainbridge1994artificial, dautenhahn2007paradigm}, which aims to advance machine comprehension and interaction within complex social contexts. The Social-IQ benchmark \cite{zadeh2019social} exemplifies this challenge through a multiple-choice question (MCQ) task, which allows the integration of video, audio, and subtitles to test models' reasoning ability regarding social dynamics. This benchmark is a valuable resource of ASI, as it mirrors the complexity of real-world social scenarios.

Social-IQ posed significant challenges for advanced VideoQA methods when it was first introduced \cite{zadeh2017tensor, lei2018tvqa, liang2018focal}: relying solely on the question and answer (BlindQA) resulted in near-random accuracy. With the ongoing advancements of Large Language Models (LLMs), recent multimodal approaches incorporating them have more than doubled this accuracy. However, as illustrated in Figure \ref{lead_in_fig} (c), these improvements are largely due to the LLMs' shortcut effect, i.e., exploiting inherent spurious correlations between questions and options, while visual context is almost entirely disregarded in the answering process. DeSIQ \cite{guo2023desiq} even found that a T5-small language model \cite{2020t5} could achieve 100\% accuracy on Social-IQ-1.0 under BlindQA, highlighting a significant language bias in Social-IQ.
Moreover, current solutions, whether based on Transformer \cite{yang2021just, pirhadi2023just}, LSTM \cite{kumar2020mcqa}, contrastive learning \cite{wilf2023face, xie2023multi}, or compositional models \cite{natu2023external}, typically rely on uniformly sampled frames that fail to provide question-relevant visual context, let alone capture the nuanced social traits embedded in the videos. 
These limitations and data biases lead models to exploit language shortcuts for selection \cite{niu2021counterfactual, cho2023generative, Lao_Pu_Liu_He_Bakker_Lew_2023}.

The consequences are profound. \textbf{i)} Such models face doubts about their ability to meet Social-IQ's goal: discerning the correct option through comprehensive reasoning about multimodal social traits. \textbf{ii)} The closed-set nature of MCQ amplifies these concerns, as it restricts the exploration and assessment of whether the selected answers truly reflect the underlying reasoning.  In contrast, open-ended answers provide deeper insights, as shown in Figure \ref{lead_in_fig} (a), where a model reliant on shortcuts reveals its ungrounded rationale. These challenges and opportunities encourage us to develop a more transparent and reliable approach to Social-IQ.

To combat these issues, in this paper, we propose a novel \textbf{V}isually \textbf{E}xplainable and \textbf{G}rounded \textbf{A}rtificial \textbf{S}ocial Intelligence (\textbf{VEGAS}) framework for Social-IQ. 
Specifically, we opt for a generative approach, as seen in recent Multimodal Large Language Models (MLLMs), to enable \textit{Explainable} open-ended responses, facilitating the probing and measuring of reasoning paths.
To deliver visually \textit{Grounded} answering, we employ a dual-pronged strategy consisting of \textbf{L}anguage \textbf{G}uided \textbf{S}ampling (\textbf{LGS}) and \textbf{G}eneralist \textbf{I}nstruction \textbf{F}ine-\textbf{T}uning (\textbf{GIFT}).
\textbf{Firstly}, the {LGS} equips the model with the ability to sample question-relevant video frames in social interactions, guided by language cues in the form of explicit descriptions, causal questioning, and nuanced differentiation. We craft the dataset and learning strategy to enable effective LGS supervision in the absence of timestamp annotations. However, the sampled frame features often exhibit disordered temporal relationships due to the temporal embedding in the pre-trained video encoder, which is optimized for uniformly spaced frames. This leads to misinterpretations of social activities, such as \textit{touching vs} \textit{hitting}, which are crucial for understanding latent social attitudes. To address this, we propose a \textbf{T}emporal \textbf{A}ttention \textbf{M}odule (\textbf{TAM}) to restore the order in these frames, ensuring coherent sequencing without a secondary encoding.
\textbf{Secondly}, the goal of {GIFT} is to learn an effective understanding of sampled visual features, which requires advanced abilities from the subsequent reasoning modules. To achieve that, we first integrate the \textbf{S}ocial \textbf{T}raits \textbf{P}rojector (\textbf{STP}) to learn transformation for fundamental emotional traits (video, image, and audio) into the language space. Following this, we perform joint fine-tuning of STP and LLM using an expansive multimodal social interaction dataset. This results in VEGAS-\textit{generalist},  an enhanced version excelling in joint reasoning, enriched social commonsense, and advanced expertise in sociology and psychology.

In this study, we prioritize the correctness of the reasoning path over mere pursuit of maximum accuracy. We uncover and assess the reasoning of open-ended answers using ChatGPT. The evident accuracy improvement of  VEGAS highlights the consistency between its reasoning and the correct selection, cf. Figure \ref{lead_in_fig} (b). Furthermore, modality ablation in MCQ reveals that VEGAS significantly suppresses language shortcuts, improving visual context utilization from -0.43\% to 9.28\%. Finally, VEGAS achieves state-of-the-art performance with a more credible and scalable implementation. 
Contributions of this paper are summarized as follows:
\begin{enumerate}		
	
	\item[$\bullet$] We for the first time introduce VEGAS, a visually explainable and grounded social intelligence model that mitigates language shortcut effect   in Social-IQ and efficiently considers visual context answering.

        \item[$\bullet$]  We propose a dual approach to enhance the relevance of video frames and improve their interpretation, ensuring accurate and visually grounded answers. 

	\item[$\bullet$] We introduce VEGAS-\textit{generalist}, a sophisticated human-like social AI that demonstrates profound understanding and analytical expertise in social dynamics.
\end{enumerate}

\section{Related Work}
\subsection{Social-IQ}

The Social-IQ-1.0 benchmark \cite{zadeh2019social} was introduced in 2019 to evaluate the social intelligence level of AI models with MCQ task. Social-IQ-2.0 soon updates the benchmark with newly annotated questions and answers. As solutions, \cite{natu2023external} incorporate external knowledge retrieved from VisualCOMET \cite{10.1007/978-3-030-58558-7_30} to augment the multimodal features with  social commonsense. MMTC-ESC \cite{xie2023multi} leverages contrastive learning with emotional cues to build cross-modal correlations of features. Just Ask Plus \cite{pirhadi2023just} uses multi-headed attention and transformer encoders to compute representations for the questions and answers, then calculates their similarity for selection. F2F-CL \cite{wilf2023face} conducts fine-grained graph contrastive learning by decomposing the social interaction according to speaking turns. Moreover, Social-IQ is also a popular benchmark in many generic video understanding models \cite{li2024llms, xu2023retrieval, fei2024video}.

Despite ongoing efforts, few studies have addressed the language shortcut issue in Social-IQ. A model-side solution \cite{gat2020removing} once proposed to use loss regularization for generic classifier debiasing, achieving a 2\% improvement on Social-IQ but lacked further analysis. DeSIQ \cite{guo2023desiq} is the only approach so far addressing language biases by empirically substituting incorrect answers with correct ones from other samples. 
In contrast, our model-side approach directly enhances visual information usage as effectively as DeSIQ but with higher accuracy.

\subsection{Multimodal Large Language Models}
Recent advancements in MLLMs have significantly enhanced video question answering (VideoQA) \cite{chen-dolan-2011-collecting,Yu2019ActivityNetQAAD,fei2024dysen,fei2024enhancing}. These models effectively integrate various modalities \cite{yu2024crema, wu2024towards,zhu2023languagebind, yu2024crema,wu24next,fei2024vitron}, such as audio, video, and depth, by projecting features from frozen encoders into the language space, leveraging them to produce natural language responses. Video-LLaMA \cite{zhang2023video} integrates visual and audio features from frozen encoders using Q-Former \cite{pmlr-v202-li23q}. Video-ChatGPT \cite{maaz2023video} uses liner layers to project temporal and spatial features extracted from videos to the LLM, and generate conversations accordingly. PG-Video-LLaVA \cite{munasinghe2023pg} strengthens the MLLM with pixel level grounding ability by introducing grounding modules like scene detector and object tracker.
Video-LLaVA \cite{lin2023video} incorporates visual encoders pre-aligned with language for unified understanding.

Although appealing, using them for social intelligence is non-trivial, as they default to uniformly sampled frames, missing critical visual details for Social-IQ.  Recent generic models that retrieve relevant video segments \cite{xu2023retrieval} or frames \cite{li2024llms} have shown some improvements on Social-IQ, but the absence of targeted social designs limits their abilities.
Despite this, the generative pipeline holds potential for bridging the gap between current research and human-like social AI \cite{chandra2022or, duenez2023social, liu2025dynamic}.

\section{Methodology}
\subsection{VEGAS Framework}
\begin{figure}[t]
	\centering 
	\includegraphics[width=0.46\textwidth]{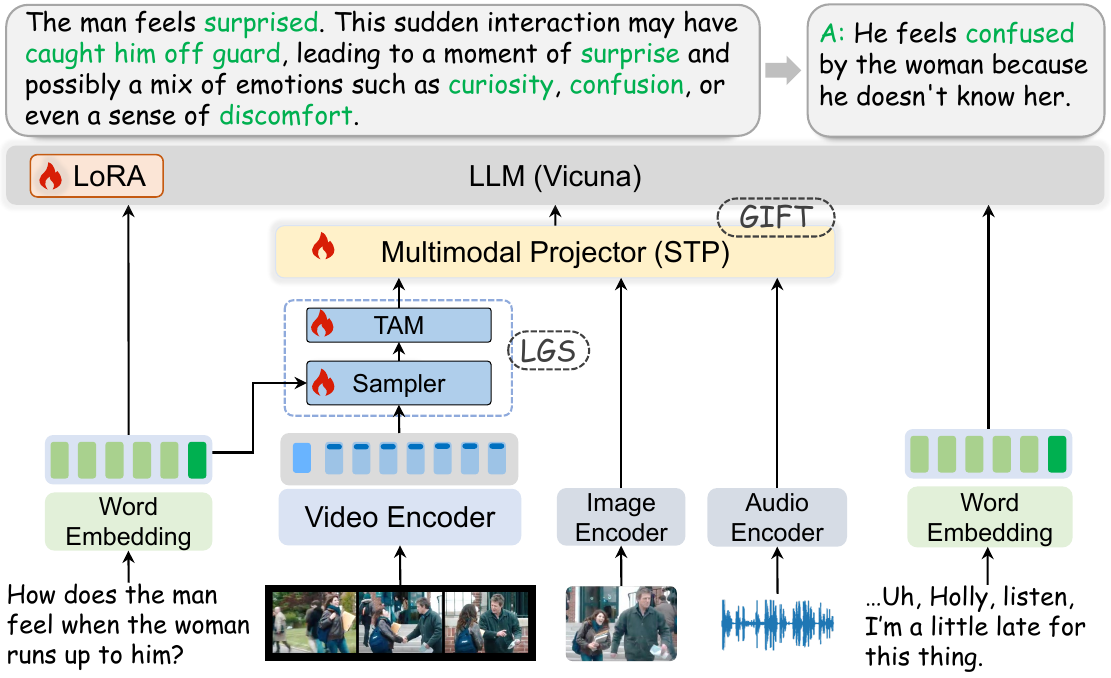}
	\caption{Architecture of VEGAS. The system encodes multimodal inputs with frozen encoders. These inputs are projected into LLM space using a trainable Multimodal Projector, enabling nuanced answer generation that captures social attitudes in interactions like emotions.} 
	\label{pipeline} 
\end{figure}

As shown in Figure \ref{pipeline}, the VEGAS framework integrates video, image, and audio encoders from LanguageBind \cite{zhu2023languagebind} to process inputs of various modalities, along with a word embedding layer for text encoding. While the video encoder is primarily used for the Social-IQ task, we include image and audio encoders for better scalability and general applicability.

First, all $n$ candidate frames are encoded by the video encoder. In the LGS, the sampler selects 
$k$ frames that align with the language hint—either a question (for inference) or its fusion with an answer (for training). The TAM then restores the temporal relationships among the sampled frames. To connect the encoders with the LLM, we use linear layers \cite{lin2023video} to build the Social Traits Projector (STP), which learns transformations of modality social traits. Based on the multimodal features and word embeddings, the LLM generates either free-form text or selected options as per user instructions. During the GIFT stage, we fine-tune the STP along with the LLM, resulting in VEGAS-\textit{generalist}.

\subsection{Language Guided Sampling (LGS)}
\subsubsection{Sampler Structure.}
Let $V$ represent the sequence of \( n \) uniformly sampled video frames, and \( Q \) denote the language hint consisting of \( m \) words.
We encode these inputs using the video and text encoders, producing $f_{v0}$ for the visual frames and $f_q$ for the text. Here, $f_{v0}$ is computed with pre-trained temporal embeddings, where each frame is associated with a \texttt{CLS} token that encapsulates its global visual feature.
To explore the causal relevance between $V$ and $Q$, we compute a contextualized representation for the visual features using the ATP transformer block \cite{buch2022revisiting} as
\begin{equation}
	f_{qv} = \text{ATP}(\mathrm{Linear\_v}(f_{v0}^{\text{CLS}}), \mathrm{Linear\_q}(f_q)).
\end{equation}
Here, $f_{v0}^{\text{CLS}}$ denotes the \texttt{CLS} token for each frame.
Based $f_{qv}$, we calculate the relevance logits as:
\begin{equation}
	logits = \mathrm{Linear\_qv}(f_{qv}) \in \mathbb{R}^{n \times d},
\end{equation}
where $d$ is the dimension of visual feature.
Then, we select the top-\( k \) frames as
\begin{equation}
	f_v = \text{Top-K}({logits}, f_{v0}) \in \mathbb{R}^{k \times d}.
	\label{eq_top_k}
\end{equation}
The Top-K function is implemented to be differentiable by introducing stochastic perturbations during training \cite{Cordonnier_2021_CVPR} for optimization. At inference, we directly select the indices of $logits$ with the top-\( k \) values as key frame indicators.

\begin{figure*}[t]
	\centering 
    \includegraphics[width=0.99\textwidth]{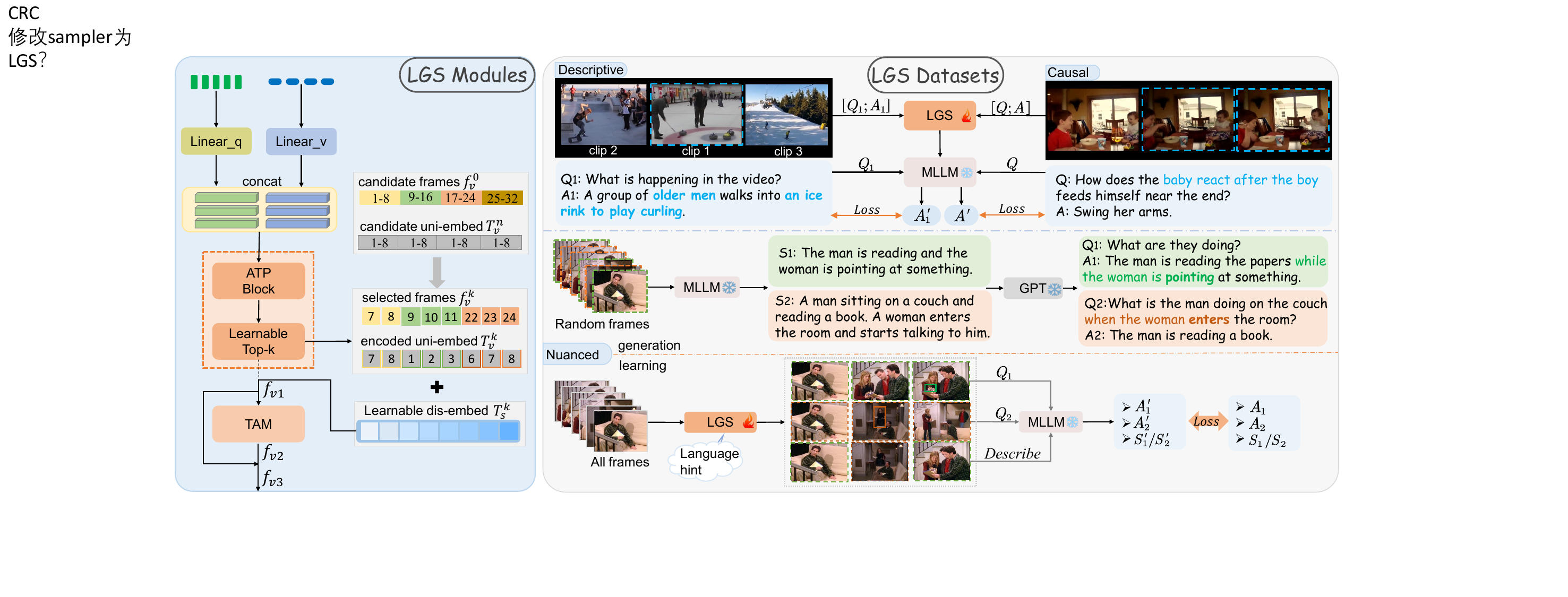}
	\caption{Left: The proposed Language Guided Sampling modules. Right: Three datasets are crafted for LGS training, incorporating descriptive, causal, and nuanced language cues. } 
	\label{sampler_training} 
\end{figure*}  

\subsubsection{Temporal Attention Module (TAM).}
The selected $k$ frame features from Eq. (\ref{eq_top_k}) are encoded with temporal embedding $T_v^k$ designed for $k$  frames (typically $k=8$). In such a process, encoding \(n\) frames \((n > k)\) requires repeating \(T_v^k\) for $n/k$ times, which disrupts temporal coherence. We introduce the TAM to restore the order of sampler output $f_v$.  First, the $f_v$ from Eq. (\ref{eq_top_k}), as illustrated in Figure \ref{sampler_training}, can be broken down as
\begin{equation}
	f_v = f_v^k + T_v^k .
\end{equation}
We additionally learn a new temporal embedding  \( T_s^k \)  for sampled frames, which is defined and initialized as
\begin{equation}
	T_s^k \in \mathbb{R}^{k \times d} \sim \mathcal{N}(0, \frac{1}{\sqrt{d}}) .
\end{equation}
Then, we construct new relationships using a CLIP attention module \cite{zhu2023languagebind} as
\begin{align}
	f_{v1} &= f_{v} + T_s^k, \\
	f_{v2} &= \text{CLIPAtt}(\text{LayerNorm}(f_{v1})), \\
	f_{v3} &= f_{v1} + f_{v2}.
\end{align}
Here, $f_{v3}$ is the final output of LGS and then used as input of STP. We use residual connections considering uniform frames may still be needed in tasks like video captioning.

\subsubsection{Sampling Data Construction.}
We craft targeted data for LGS learning due to the lack of timestamp annotations. We aim for three kinds of sampling ability as illustrated in Figure \ref{sampler_training}.
\textbf{Firstly}, the basic localization ability based on \textit{explicit description}. We create a composite video \( V \) by randomly selecting, shuffling, and merging three video clips from the Video-ChatGPT dataset \cite{maaz2023video}. The training sample is then defined as \( (V, Q_1, A_1) \), where the QA pair from clip1 is used for video captioning training. This setup compels the sampler to accurately locate frames in clip1 when provided with the language hints \( Q_1 \) and \( A_1 \).
\textbf{Secondly}, the ability to capture frames that have \textit{causal relations} with language hints. For this, we employ the QA samples from the temporal subset of Next-QA \cite{xiao2021next}.

\textbf{Thirdly}, the ability to distinguish key frames based on \textit{nuanced details} in less varying social interactions.We use the TVQA dataset \cite{lei2018tvqa} for its rich social dynamics. 
We use a self-refinement strategy to enforce the model to recall and locate these dynamics according to its own memory. For each video, we begin by creating pseudo samples through two rounds of random sampling ($i \in \{1,2\}$), each selecting \( k \) frames from the video. We then use the baseline model to generate captions \( S_1 \) and \( S_2 \) for the two clips, which serve as the memory. Next, we prompt ChatGPT to craft distinctive QA pairs based on the captions. Finally, we construct two samples of the same video as
\begin{equation}
	\begin{aligned}
		& \text{Sample}_i = \left\langle V, Q_i, A_i, S_i\right\rangle , i\in \{1, 2\} \,.
	\end{aligned}
\end{equation}
This design ensures that \(Q_i\) can only be correctly answered with  \(A_i\), with temporal moment in $V$ captured by \(S_i\).

\subsection{Training Pipeline}
\subsubsection{LGS Training.}
To effectively train the LGS modules, we employ QA pairs as language hints of sampler for the descriptive and causal samples generated above. The predicted free-form answers $A_1'$ and $A'$ are supervised with their ground truths.
For the generated pseudo data, each sample is utilized in two ways, as depicted in Figure \ref{sampler_training}. \textbf{i)} The QA task is formulated as
\begin{equation}
	A_i' = \text{LLM}({Q_i}, \mathcal{S}({V}, {Q}_i, {A}_i)), i\in \{1, 2\},
\end{equation}
where the sampler \(\mathcal{S}(\cdot)\) generates key frame features for \({V}\) conditioned on $Q_i$ and $A_i$.
\textbf{ii)} The video captioning task is conducted as
\begin{equation}
	S_i' = \text{LLM}({Q}, \mathcal{S}({V}, {S}_i, {A}_i)), i\in \{1, 2\},
\end{equation}
where $Q=\textit{``Describe the video.''}$. The model predictions are supervised with the CrossEntropyLoss function.

\subsubsection{GIFT 1: STP Emotion Transformation Learning. }
We initialize the STP module using the linear visual projector from Video-LLaVA \cite{lin2023video}, which ensures optimal alignment between visual and language modalities through captioning tasks. For social intelligence purposes, we customize STP 
by multimodal-language transformation learning using emotion recognition as the proxy task. The process is formulated as optimizing $p(E\ |\ \text{LLM(STP}(Z)))$, where $Z$ represents the encoded multimodal features and $E$ denotes the predicted emotional category. In this process, Mel-spectrogram features are extracted from audio tracks and encoded with a Vision Transformer (ViT). We also incorporate audio captioning data to augment the model’s understanding of the ongoing audio events.

\subsubsection{GIFT 2: STP\&LLM Joint Representation Learning.}
We fine-tune the STP and the LLM together on an extensive multimodal social interaction dataset, which is helpful for joint reasoning and human-aligned answering. By ``joint", we refer to both the model and data aspects, allowing a sample to contain multiple modalities instead of only one.
Specifically, we process multimodal input \texttt{<mm>} concurrently by concatenating visual features with audio features (if any). For subtitles, we treat them as an individual modality in training to differentiate the primary user instruction (question) from the dialog content. This also helps in understanding ongoing social interactions through conversations. To enhance robustness, we organize modalities of each training sample using two sequences: \texttt{<mm><subtitle><question>} and \texttt{<question><mm><subtitle>}.

\section{Experiments}

\subsection{Settings}

\subsubsection{Data Details.}
In this study, we report results on Social-IQ-2.0 and leverage various datasets and their transformations in training as Table \ref{tab_datasets} shows. For the LGS, we craft data based on TVQA \cite{lei2018tvqa}, Next-QA \cite{xiao2021next}, and Video-ChatGPT \cite{maaz2023video}. For the STP, we use RAVDESS \cite{livingstone2018ryerson}, AudioCaps \cite{kim2019audiocaps}, CMU-MOSEI \cite{zadeh2016multimodal}, and Expression in-the-Wild (ExpW) \cite{zhang2018facial}.
For VEGAS-\textit{generalist}, we integrate TVQA and CMU-MOSEI for multimodal joint training. We incorporate expert insights distilled by ChatGPT from Social-IQ data  \cite{zadeh2019social} to provide in-depth analysis. We also use a portion of Video-ChatGPT data to mitigate catastrophic forgetting. We use the original Social-IQ in MCQ experiments. The combined dataset totals approximately 240,000 samples. 
\begin{table}[h]
	\fontsize{8}{8}\selectfont
	\setlength{\tabcolsep}{9pt}
	\centering
	\setlength{\tabcolsep}{5pt} 
	\begin{tabular}{l | c | c | c}
		\toprule
		\textbf{Dataset} & \textbf{Sampling} & \textbf{STP} & \textbf{Joint} \\
		\midrule
		TVQA & \includegraphics[height=8pt]{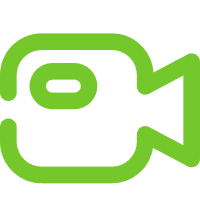}  & & \includegraphics[height=8pt]{figures/video.png}
		\includegraphics[height=8pt]{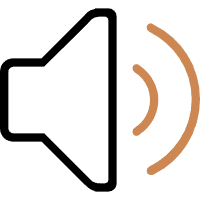}
		\includegraphics[height=8pt]{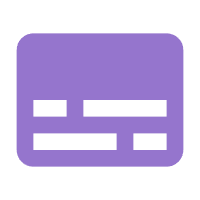}\\
		Next-QA & \includegraphics[height=8pt]{figures/video.png}  & & \\
		Video-ChatGPT  & \includegraphics[height=8pt]{figures/video.png}  & & \includegraphics[height=8pt]{figures/video.png} \\
		RAVDESS &  & \includegraphics[height=8pt]{figures/audio.png} & \\
		AudioCaps &  & \includegraphics[height=8pt]{figures/audio.png} & \\
		CMU-MOSEI &  & \includegraphics[height=8pt]{figures/video.png} \includegraphics[height=8pt]{figures/audio.png} & 
		\includegraphics[height=8pt]{figures/video.png}
		\includegraphics[height=8pt]{figures/audio.png}
		\includegraphics[height=8pt]{figures/subtitle.png} \\
		ExpW &  & \includegraphics[height=8pt]{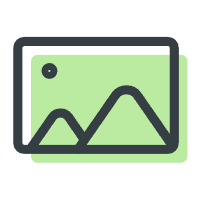} & \\
		Social-IQ-\textit{expert}\&MCQ & & &  
		\includegraphics[height=8pt]{figures/video.png}
		\includegraphics[height=8pt]{figures/subtitle.png} \\
		\midrule
		Data size &  33k & 165k & 43k \\
		\bottomrule
	\end{tabular}
	\caption{Datasets used at different training stages.}
	\label{tab_datasets}
\end{table}

\begin{figure*}[t]
	\centering 
	\includegraphics[width=0.96\textwidth]{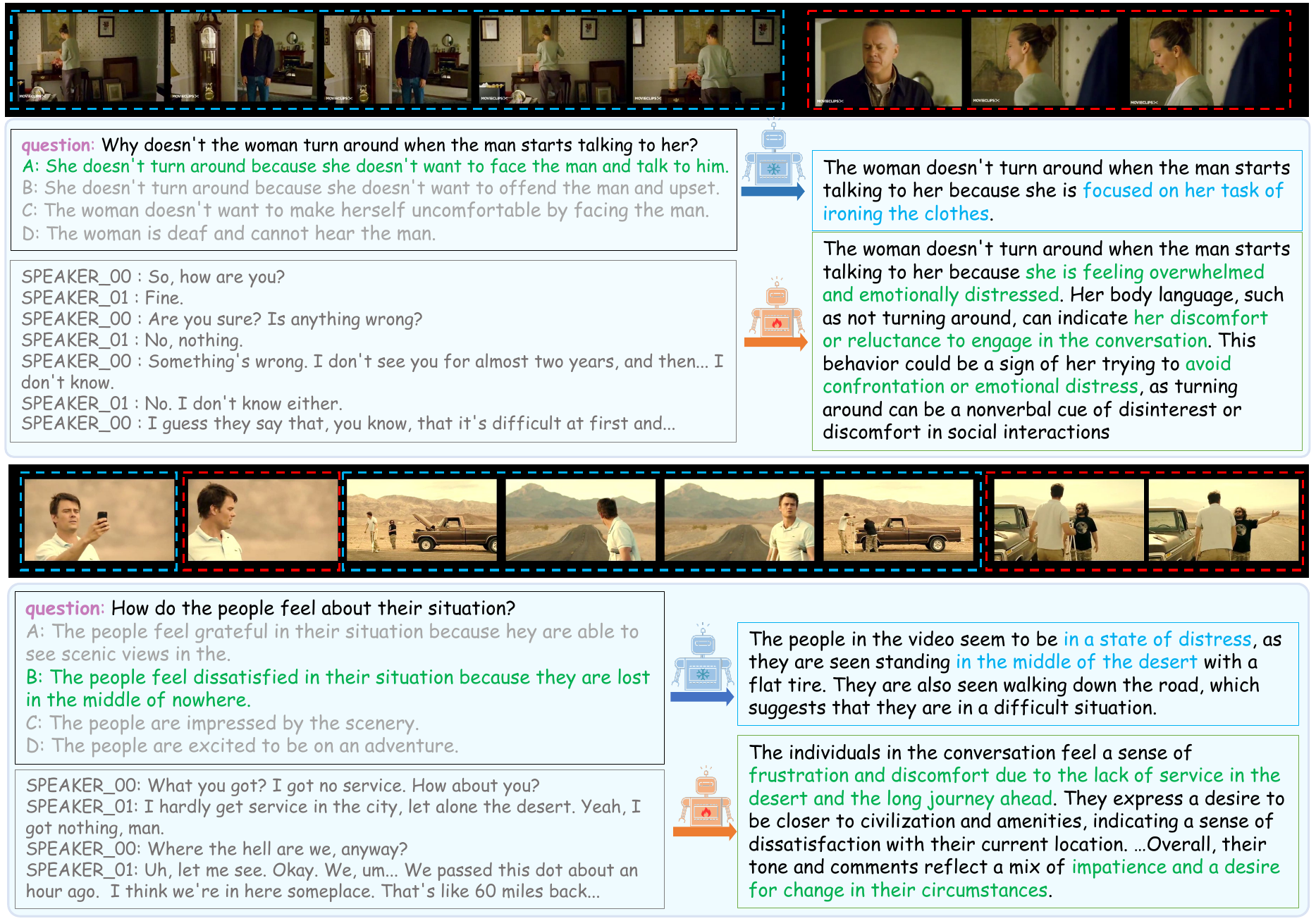}
	\caption{Open-ended QA examples of VEGAS (\textcolor{blue}{blue} arrows) and VEGAS-\textit{generalist}  (\textcolor{orange}{orange} arrows) using video alone and video with subtitles, respectively.  } 
	\label{general_qualitative} 
\end{figure*}  

\subsubsection{Training Details.}
For the sampler,  we set $n=32$ and $k=8$, and encode language hints with the text encoder from CLIP ViT-B/32 \cite{Radford2021LearningTV}. We initiate the LGS from scratch and train the sampling process with a learning rate of 2e-4. The STP module is pre-trained with a learning rate of 1e-6. 
For the joint tuning of the STP and LLM, we set their learning rates to 2e-5 and 2e-4, respectively and train for three epochs. The Vicuna-7b LLM \cite{chiang2023vicuna} is fine-tuned using Low-Rank Adaptation (LoRA) with \(r=128\) and \(\alpha=256\). Note that the joint tuning is performed for both VEGAS-\textit{generalist} in open-ended QA and VEGAS in supervised MCQ, but on different datasets. All training is conducted on 4 A100 40G GPUs with a batch size of 64.  All training proceeds for one epoch except for supervised MCQ, which is trained for three epochs.

\subsubsection{Metrics.}
Following Video-ChatGPT \cite{maaz2023video}, we use GPT-3.5-turbo to assess Accuracy (\%) and Score (1-5) for open-ended answers. Accuracy determines if the answer matches the correct option, while the Score measures how closely they are aligned. All four candidate options are included in the evaluation for rigor. For MCQ, Accuracy is calculated directly by literal matching. We use the first letter of  \textbf{Q}uestion, \textbf{A}nswers, \textbf{V}ideo, and \textbf{S}ubtitles to denote each modality, respectively.

\subsection{Open-Ended QA}

\subsubsection{Zero-shot Results.}
We start by probing the correctness of the reasoning path in the answers.  Table \ref{zero_shot} presents results comparison with strong MLLM baselines, and module ablations of our proposed designs.
By zero-shot, we refer to the VEGAS model with LGS rather than the fine-tuned VEGAS-\textit{generalist}.

The ablations in the \textbf{QV} setting demonstrate the effectiveness of the proposed sampling strategy and the temporal attention. We find that using fewer candidate frames ($n=16$) improves accuracy but harms the consistency score. This might indicate that the generated answers only roughly align with the correct option, but fail to cover details or rationales. The TAM addresses this problem by reconstructing temporal relationships when working with more candidate frames. Interestingly, even though the VEGAS model primarily affects vision branch, the improved \textbf{QVS} metric (49.5\% \textit{vs} 51.2\%) indicates that better visual features can enhance the model's understanding of the conversation in subtitles.

\begin{table}[htbp]
	\fontsize{8.5}{8.5}\selectfont
	\setlength{\tabcolsep}{9pt}
	\centering
	\setlength{\tabcolsep}{1pt} 
	\begin{tabular}{lcccc}
		\toprule
		\multirow{2}{*}{\textbf{Model}} & \multicolumn{2}{c}{\textbf{QV}} & \multicolumn{2}{c}{\textbf{QVS}}  \\
		\cmidrule(lr){2-3} \cmidrule(lr){4-5}
		& \textbf{Score} & \textbf{Accuracy} & \textbf{Score} & \textbf{Accuracy}  \\
		\midrule
		Video-LLAMA & 1.6 & 36.1 & 1.7 & 37.9 \\
		Video-ChatGPT & 3.4 & 43.5 & 3.4 & 49.2 \\
		PG-Video-LLaVA & 3.4 & 42.8 & 3.5 & 48.5 \\
		Video-LLaVA & 3.4 & 42.2 & 3.4 & 49.5 \\
		\midrule
		\textcolor{lightgray}{VEGAS $_\textit{n=16}$ $_\text{w/o TAM}$ }  & \textcolor{gray}{2.7} & \textcolor{lightgray}{43.6} & \textcolor{gray}{-} & \textcolor{gray}{-} \\
		\textcolor{gray}{VEGAS $_\textit{n=16}$}  & \textcolor{gray}{2.8} & \textcolor{gray}{46.1} & \textcolor{gray}{-} & \textcolor{gray}{-} \\
		VEGAS  $_\textit{n=32}$  $_\text{w/o TAM}$ & 3.2 & 46.1 & 3.3 & 49.0 \\
		
		VEGAS  $_\textit{n=32}$  & 3.4 & 46.1 & 3.5 & 51.2 \\
		VEGAS-\textit{generalist} $_\text{w/o STP}$  & 3.1 & 48.4 & 3.6 & 51.5 \\
		VEGAS-\textit{generalist}   & 3.4 & 48.5 & 3.9 & 54.9 \\
		\bottomrule
	\end{tabular}
	\caption{Modality and module ablation results in the open-ended setting.}
	\label{zero_shot}
\end{table}

\subsubsection{VEGAS-\textit{generalist}. }
The results in Table \ref{zero_shot} highlight the modular effectiveness of GIFT and its learning strategy. Similar to TAM in VEGAS, the STP module significantly improves VEGAS-\textit{generalist}, mitigating score drops from newly introduced designs. The STP shows notable gains in both Score and Accuracy across QV and QVS settings. The improvement in QVS further supports the beneficial interactions between modalities. Note that, despite leveraging Social-IQ expertise in GIFT, we avoid using answers as labels directly for better generalization ability.

Figure \ref{general_qualitative} shows examples comparing answers from frozen VEGAS and tuned VEGAS-\textit{generalist}. VEGAS accurately identifies relevant frames and provides correct visual evidence (e.g., “\textit{ironing the clothes}”), but struggles with in-depth analysis. As expected, VEGAS-\textit{generalist} provides responses well-aligned with ground truths and enriched with expert analysis, demonstrating a deeper understanding.

\subsection{Multi-Choice QA}
\subsubsection{Modality Ablation.}
We first perform modality ablation studies in supervised MCQ. Table \ref{modal_ablation} shows that while DeSIQ improves performance in setting A, its advantage wanes in setting QA due to unresolved data bias. Additionally, it has difficulty understanding conversations in subtitles.
In contrast, VEGAS mitigates shortcut effect in both A and QA settings while improving subtitle comprehension. More importantly, it significantly enhances the use of visual information, with a notable accuracy increase of 9.28\%.

We also conduct full-parameter fine-tuning of LLM to explore potential improvements over LoRA tuning. Unexpectedly, the baseline method performs worse under this setting, likely due to the disruption of strong prior knowledge. Contrastively, despite a decrease in maximum accuracy for VEGAS-\textit{full}, it surprisingly demonstrates significant improvements in reducing language shortcuts and enhancing the visual modality contribution by 26.21\%.
This evident advantage proves that, despite the overfitting risk associated with full-parameter fine-tuning, our approach avoids relying on spurious correlations in the language input.
In subsequent experiments, we continue with the LoRA version of VEGAS due to its comprehensive and balanced performance.
\begin{table}[t]
	\fontsize{8}{8}\selectfont
	\setlength{\tabcolsep}{9pt}
	\centering
	\setlength{\tabcolsep}{5pt}
	\begin{tabular}{lccll}
		\toprule
		\textbf{Model}  & \textbf{A}$\downarrow$ & \textbf{QA} $\downarrow$ & \textbf{QAV} & \textbf{QAVS}  \\ 
		\midrule
		DeSIQ$^{*}$ & 63.35 & 64.63 & 62.28 $_{(-2.35)}$ & \ \ \ \ -   \\ 
		DeSIQ  & 28.07 & 57.23 & 68.93$_{(+11.7)}$ & 37.72  {\fontsize{8}{12}\selectfont \faVolumeUp} \\ 
		\midrule
		Video-LLaVA  & 69.57 & 77.77  & 77.34 $_{(-0.43)}$ & 79.17  \\ 
		VEGAS  & 57.00 & 66.23 & 75.51 $_{(+9.28)}$ & 80.90   \\ 
		\midrule
		Video-LLaVA-\textit{full} & 66.01   &73.46 &74.75$_{(+1.29)}$ & 75.51 \\
		VEGAS-\textit{full} &30.53  & 47.46& 73.67$_{(+26.21)}$& 76.37  \\
		\bottomrule
	\end{tabular}
	\caption{Modality ablation results in supervised MCQ. {\fontsize{8}{12}\selectfont \faVolumeUp} denotes that the audio modality is used along with the subtitles. * denotes baseline of the method.}
	\label{modal_ablation}
\end{table}

\begin{table}[htbp]
	\fontsize{8}{8}\selectfont
	\setlength{\tabcolsep}{9pt}
	\centering
	\begin{tabular}{l|lcc}
		\toprule
		\textbf{Mode} & \textbf{Model}  & \textbf{Setting} & \textbf{Accuracy} \\ 
		\midrule
		\multirow{8}{*}{\rotatebox{90}{Supervised}} & Just-Ask & MC & 52.12 \\
		& Just-Ask-Plus & MC & 53.35 \\
		& DeSIQ$^{*}$ & MC & 64.63 \\
		& DeSIQ  & MC & \ \ \ \ 74.13 {\fontsize{8}{12}\selectfont \faVolumeUp} \\
		& MMTC-ESC* & MC & 74.91 \\
		& MMTC-ESC & MC & 75.94 \\
		& Video-LLaVA & MC & 79.17 \\
		& VEGAS & MC & 80.90 \\
		\midrule
		\multirow{6}{*}{\rotatebox{90}{Zero-shot}} & R-VLM  & Unknown & 63.7 \\
		& IVA & Unknown & 68.0 \\
		& Video-LLaVA  & MC & 60.6 \\
		& Video-LLaVA & OE & 52.5 \\
		& VEGAS & MC & 60.0 \\
		& VEGAS & OE & 66.0 \\
		\bottomrule
	\end{tabular}
	\caption{Maximum accuracy comparison under supervised (upper) and zero-shot (lower) settings. {\fontsize{8}{12}\selectfont \faVolumeUp} denotes that the audio is used as an auxiliary  modality.}
	\label{max_acc}
\end{table}

\subsubsection{Maximum Accuracy Comparison.}
The upper part of Table \ref{max_acc} compares maximum accuracy across models. DeSIQ shows better performance when using audio rather than subtitles as an auxiliary modality. Overall, the proposed VEGAS model demonstrates state-of-the-art performance. There are some generative vision-language models \cite{xu2023retrieval, li2024llms} also reported zero-shot \textbf{binary} Accuracy evaluated with ChatGPT on Social-IQ, which means they only compare the prediction with the correct answer. It is also unclear whether they handle open-ended or closed-set QA, the latter being defined by provided answer options.
Therefore, we compare under both scenarios in the lower part of Table \ref{max_acc}. Note that IVA was jointly trained with 34k NEXT-QA samples in addition to 136k instruction-tuning data. In contrast, our VEGAS model, trained on only 33k samples designed for the sampler, achieves 66.0\% accuracy, closely matching IVA's 68.0\%.

\subsection{Exploring Emotion Understanding Ability}
\subsubsection{Emotion Recognition.}
Emotions are crucial indicators of people's attitudes during social interactions. We use the IEMOCAP \cite{busso2008iemocap} dataset for multimodal emotion recognition validation, which includes emotional conversations from actors in both scripted and spontaneous spoken scenarios. 
As shown in Table \ref{emotion_recognition_results}, baseline models display varying performance, likely due to characteristics of their instruction tuning datasets. For instance, Video-ChatGPT is trained on ActivityNet-200 \cite{Heilbron_2015_CVPR} with a stronger focus on human activities. VEGAS-\textit{generalist} achieves superior performance: 25.5\% with video alone, 35.8\% with video and subtitles, and 37.4\% when incorporating audio additionally.

\subsubsection{Emotional Video Captioning.}
When observing social events, we humans naturally perceive the underlying sentiments and the nuanced emotions at play. In this part, we investigate such ability on a random subset of EmVidCap dataset \cite{9352546}. The dataset bridges visual content and linguistic sentences by combining factual and emotional elements in captions. We use uniformly sampled frames as input since a captioning task often requires full-length video understanding.

\begin{figure}[htbp]
	\centering 
	\includegraphics[width=0.46\textwidth]{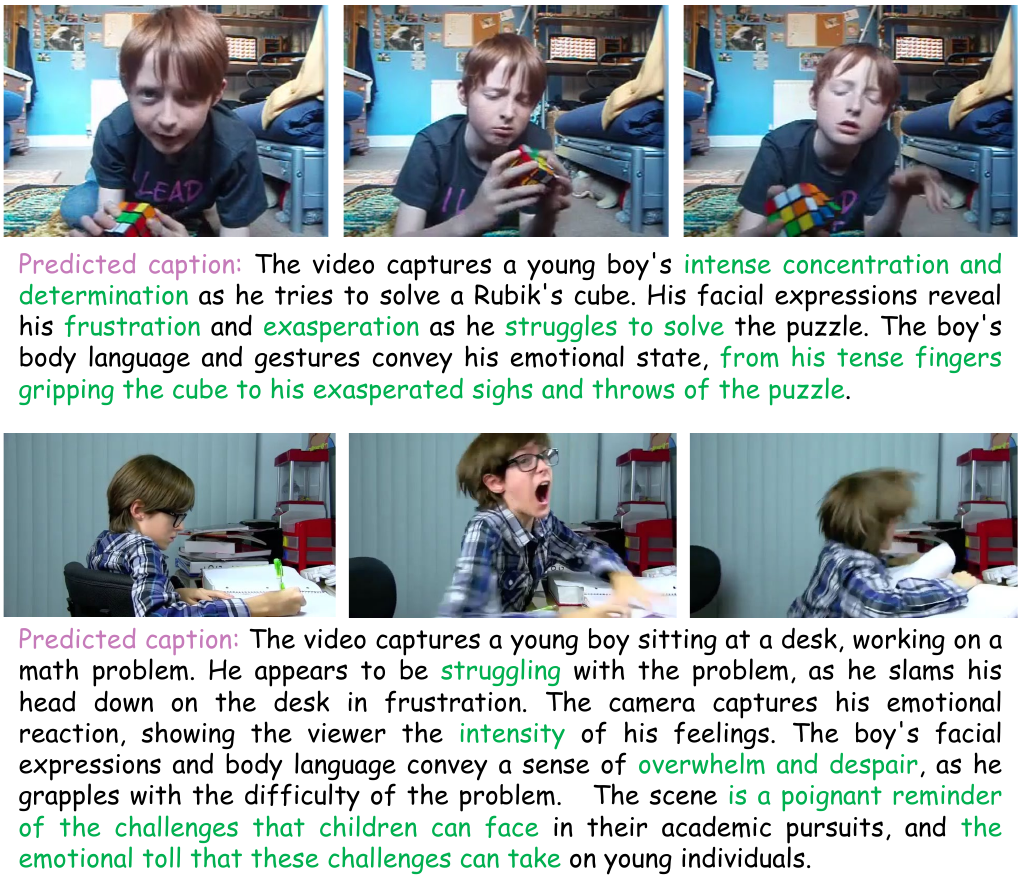}
	\caption{Video captioning examples from EmVidCap.  } 
	\label{fig_emo_vid_cap} 
\end{figure}

The lower part of Table \ref{emotion_recognition_results} presents scores reported by GPT-3.5-turbo, which was prompted to focus on emotional consistency.  Despite the zero-shot setting, VEGAS-\textit{generalist} gains evident advantages over generic MLLM baselines.
Examples in Figure \ref{fig_emo_vid_cap} further demonstrate human-like reasoning and analyzing processes, underscoring the intelligence level and broader implications of our approach.

\begin{table}[t]
	\fontsize{9}{9}\selectfont
	\setlength{\tabcolsep}{9pt}
	\centering
		\begin{tabular}{@{}l p{1.2cm} c c@{}}
			\toprule
			\textbf{Model} & \textbf{V} & \textbf{V+Sub} & \textbf{V+Sub+Aud} \\
			\midrule
			Video-LLAMA & 15.3 & 17.8 & 17.1 \\
			Video-ChatGPT & 22.2 & 25.0 & - \\
			PG-Video-LLaVA & 7.6 & 18.9 & - \\
			Video-LLaVA & 16.4 & 16.6 & - \\
			VEGAS-\textit{generalist} & 25.5 & 35.8 & 37.4 \\
			\midrule
			Video-LLAMA & 0.5 &0.5  &0.4  \\
			Video-ChatGPT & 1.1 & 1.3 & - \\
			PG-Video-LLaVA & 0.9 & 1.2 & - \\
			{Video-LLaVA} & 1.5 & 2.0 & - \\
			{VEGAS-\textit{generalist}}& 1.7 &2.3& 1.7 \\
			\bottomrule
		\end{tabular}
		\caption{Accuracy of emotion recognition on IEMOCAP (upper), and ChatGPT score on EmVidCap (lower). }
		\label{emotion_recognition_results}
	\end{table}
	
	\section{Conclusion}
	In this study, VEGAS is introduced to address the trust crisis in Social-IQ, where high selection accuracy often stems from significant language shortcuts. VEGAS utilizes a modality scalable, generative, multimodal large language model (MLLM) to deliver open-ended answers that reveal the reasoning behind selections. We first incorporate a novel Language Guided Sampling (LGS) technique to extract question-relevant visual features. Then we employ a specialized Generalist Instruction Fine-Tuning (GIFT) to produce VEGAS-\textit{generalist} to interpret social traits from them, which excels in broader social reasoning with expert-level analytical capabilities. Extensive evaluations demonstrate that VEGAS significantly enhances the integration of visual context, ensuring that it plays a pivotal role in reasoning, while effectively mitigating reliance on language shortcuts. Notably, VEGAS-\textit{generalist} excels in social understanding with expertise in psychology and sociology, positioning it as an advancing human-like social AI.

\section{Acknowledgments}
This work was supported by the National Natural Science Foundation of China (62171325). The numerical calculations in this paper have been done on the supercomputing system in the Supercomputing Center of Wuhan University.

\bibliography{aaai25}

\end{document}